\documentclass[10pt,twocolumn,letterpaper]{article}

\usepackage[margin=0.9in]{geometry}
\usepackage{times}
\usepackage{epsfig}
\usepackage{graphicx}
\usepackage{amsmath}
\usepackage{amssymb}

\usepackage{subcaption}

\usepackage[pagebackref=true,breaklinks=true,letterpaper=true,colorlinks,bookmarks=false]{hyperref}

\begin{document}

\title{Single Image Reflection Removal Using Deep Encoder-Decoder Network}

\author{Zhixiang Chi$^{1}$, Xiaolin Wu$^{2}$, Xiao Shu$^{2}$ and Jinjin Gu$^{3}$\\
 $^{1}$McMaster University\\
 $^{2}$Shanghai Jiao Tong University\\
 $^{3}$The Chinese University of Hong Kong, Shenzhen\\
 {\tt\small chiz@mcmaster.ca}, {\tt\small \{xwu510,shux\}@sjtu.edu.cn}, {\tt\small 115010148@link.cuhk.edu.cn}
}
\date{}
\maketitle

\begin{abstract}
  Image of a scene captured through a piece of transparent and
  reflective material, such as glass, is often spoiled by a
  superimposed layer of reflection image.  While separating the
  reflection from a familiar object in an image is mentally not
  difficult for humans, it is a challenging, ill-posed problem in
  computer vision.  In this paper, we propose a novel deep
  convolutional encoder-decoder method to remove the objectionable
  reflection by learning a map between image pairs with and without
  reflection.  For training the neural network, we model the physical
  formation of reflections in images and synthesize a large number of
  photo-realistic reflection-tainted images from reflection-free
  images collected online.  Extensive experimental results show that,
  although the neural network learns only from synthetic data, the
  proposed method is effective on real-world images, and it
  significantly outperforms the other tested state-of-the-art
  techniques.
\end{abstract}

\section{Introduction}

Photographing a scene behind a transparent medium, most commonly
glasses, tends to be interfered by the reflections of the objects on
the side of the camera.  The intended reflection-free image, which we
call the transmission image $T$, becomes intertwined with the
reflection image $R$, and is consequently recorded as a mixture image
$I$.  The reflections cause annoying image degradations of arguably
the worst kind and make many computer vision tasks, such as
segmentation, classification, recognition, etc., very difficult if not
impossible.  For a range of important applications, the separation and
removal of reflection image $R$ from the acquired mixture image $I$ is
a challenging image restoration task out of necessity.

\begin{figure}
  \centering
  \begin{subfigure}{0.49\linewidth}
    \centering
    \includegraphics[width=0.98\linewidth]{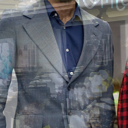}
  \end{subfigure}
  \begin{subfigure}{0.49\linewidth}
    \centering
    \includegraphics[width=0.98\linewidth]{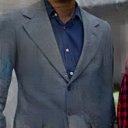}
  \end{subfigure}
  \caption{The proposed reflection removal technique separates
    reflections from a given reflection-interfered image (left) and
    outputs a clean image (right) without the reflection artifacts.}
  \label{fig:exampleIntroFirst}
\end{figure}

A widely adopted and satisfactory model for the formation of the
mixture image is
\begin{equation}
  \label{eq:itr} I = \alpha T + \beta R + n.
\end{equation}
where $n$ is the noise term, $\alpha$ and $\beta$ are the
transmittance and reflection rate of the glass, respectively; they
determine the mixing weights of the two component images.  Compared
with other image restoration tasks, such as denoising,
superresolution, deblurring, etc., reflection removal is far more
difficult.  The underlying inverse problem is one of blind source
separation and it is more severely underdetermined as there are not
one but two unknown images $T$ and $R$ that need to be estimated from
the observed image $I$.  Adding to the level of difficulty is that
both component signals $T$ and $R$ are natural images of similar
statistics.

Many researchers have taken on the technical challenge of reflection
removal and proposed a number of solutions for the problem.  But the
current state of the art is still quite limited in terms of the
performance, robustness and generality.  One approach is the use of
specially designed optical devices, such as polarizing filters, to
obtain a series of perfectly aligned images with different levels of
reflections for layer separation \cite{kong2014physically}.  Although
such optical devices make the reflection removal problem easier to
tackle, they incur additional hardware costs, reduce light influx, and
have limited scope of applications.  Thus, many techniques use
multiple images of the targeted scene taken from slightly different
viewing positions instead to get varied reflections
\cite{gai2012blind, li2013exploiting}.  However, as these techniques
require accurate image registration, they are only applicable when the
imaged objects are relatively flat and not in motion.

Ideally, a reflection removal algorithm should work with a single
mixture image, albeit a daunting task.  Some attempts have been
reported in the literature on single-image reflection removal
\cite{levin2004separating, fan2017generic}.  In these papers, the
authors adopted the image formation model of Eq.~\eqref{eq:itr} and
formulated the problem as the decomposition of the observed mixture
image $I$ into two components $T$ and $R$ of different
characteristics.  In order to separate the reflection from the
transmission, they all made some explicit assumptions about the
reflection image $R$ so it can be distinguished from the transmission
image $T$.  For instance, Shih et~al. proposed to use the double image
caused by the two surfaces of the glass to identify reflection
\cite{shih2015reflection}.  But unless the camera is close to the
glass, the double image effect is too insignificant to be useful.
Other techniques also try to exploit the smoothness and sparsity of
the reflection layer \cite{li2014single}.  However, transmission image
can have smooth and sparse regions as well, making these priors
non-discriminative.

Apparently, the task of single image reflection removal has been
greatly hindered by the inability of conventional statistical models
to separate reflection and transmission images.  We humans can, on the
other hand, mentally separate the two images although being visually
disturbed.  The difference lies in that humans can perform the
separation task largely relying on the coherence of high-level
semantics of the two images, while existing models cannot.  This
suggests that machine learning is a sensible and promising strategy
for overcoming the persistent difficulties in removing reflections
based on a single mixture image.  Furthermore, the machine learning
approach, with properly chosen training data, can also circumvent the
obstacle of grossly ill conditioning in solving the problem directly
using the image formation model of Eq.~\eqref{eq:itr}, in which the
number of unknowns greatly exceeds the number of equations.

In this paper, instead of using an explicit model like most existing
techniques, we propose a data-driven approach based on deep
convolutional neural network (CNN) for removing reflections in a
single image as shown in \figurename~\ref{fig:exampleIntroFirst}.
Similar to many CNN based image restoration techniques for problems
like inpainting, denoising and superresolution, the proposed approach
recovers a reflection-free image from a given image by learning an
end-to-end mapping of image pairs with and without reflections.  To
fully exploit the fact that the reflection image, the addictive
``noise'' to be removed, is also a natural image as discussed
previously, we design a novel three-stage deep encoder-decoder network
that first estimates the reflection layer and then reconstructs a high
quality transmission layer based on the estimated reflection and
perceptual merits.  Due to the difficulty of obtaining a sufficiently
large set of real image pairs for training the neural network, we
carefully model the physical formation of reflection and synthesize a
large number of photo-realistic reflection-tainted images from
reflection-free images collected online.  Extensive experimental
results show that the neural network can generalize well using only
synthesized training data and significantly outperform other tested
techniques for real-world images.

The remainder of the paper is organized as follows.  Section 2
provides an overview of related work in the literature.  In Section 3
we introduce our method for synthesizing training data, and in Section
4 we present in detail our proposed method.  Section 5 shows the
experiment evaluations of the proposed method with both synthetic and
real-world images.  Finally, Section 6 concludes.

\section{Related Work}

Many existing reflection removal methods rely on two or more input
images of the same scene with different reflections to estimate the
transmission layer.  To obtain such images of varied reflections,
several photography techniques can be used.  Some reflection removal
methods employ polarizing filter to manipulate the level of the
reflections \cite{ohnishi1996separating, farid1999separating,
  sarel2004separating}.  The physically-based method proposed by
Schechner et~al. shows the advantages of using orthogonal polarized
input images \cite{schechner2000polarization}.  Kong et~al. further
improve this idea by exploiting the spatial properties of polarization
\cite{kong2014physically}.  Similar to polarizing filtering, flash
lighting \cite{feris2004specular, agrawal2005removing} and defocus
blurring \cite{schechner2000blind, schechner2000separation} can help
generate images with varied reflections without moving the camera as
well.  Keeping the camera relatively stationary is crucial to the
efficacy of these device-based reflection removal techniques, as if
the objects behind the glass are also not in motion, the only changing
components among the input images are the reflections while the
transmission layer is invariant.

There are also many multi-image techniques that exploit the motion of
the camera as a cue for reflection removal.  These techniques first
align the objects in a series of images taken from slightly different
viewing positions and then separate the invariant layer as the
reflection-free image \cite{szeliski2000layer, gai2012blind}.  To
align images interfered by reflections, Tsin
et~at. \cite{tsin2006stereo} and Sinha et~al. \cite{sinha2012image}
use efficient stereo matching algorithms.  Guo et~al. exploit the
sparsity and independence of the transmission and reflection layers to
improve the robustness of image alignment \cite{guo2014robust}.
Off-the-shelf optical flow algorithms are also employed for aligning
images \cite{li2013exploiting, han2017reflection}.  With motion
smoothness constraints, optical flow techniques can be more accurate
and robust for the layer separation task \cite{xue2015computational,
  yang2016robust}.  For the cases where the camera is stationary while
the objects are moving, the reflection layer is relatively static and
must be handled differently \cite{sarel2005separating,
  simon2015reflection}.

If only one image of the scene is given, which is the case tackled by
this paper, the task of reflection removal becomes much more
challenging.  Only a few single image reflection removal techniques
have been reported in the literature.  Many of these techniques still
rely on some extra information provided by light field camera
\cite{chandramouli2016convnet, ni2017reflection} or the user
\cite{levin2007user, yeung2008extracting}.  One of the first attempts
to solve the problem without any user assistance is
\cite{levin2004separating}, which minimizes the total amount of edges
and corners in the two decomposed layers of the input image.  Akashi
et~al. \cite{akashi2014separation} employ sparse non-negative matrix
factorization to separate the reflection layer without a explicit
smoothness prior.  The work of Li and Brown \cite{li2014single}
assumes that the reflection layer is smoother than the transmission
layer due to defocus blur and hence has a short tail gradient
distribution.  With a similar smoothness assumption for the reflection
layer, Fan et~al. \cite{fan2017generic} use two cascaded CNN networks
to reconstruct a reflection reduced image from the edges of the input
image.  Arvanitopoulos et~al. \cite{arvanitopoulos2017single}
formulate the reflection suppression problem as an optimization
problem with a Laplacian data fidelity term and a total variation
term.  Wan et~al. \cite{wan2017sparsity} combine the sparsity prior
and nonlocal prior of image patches in both the transmission and
reflection layers together.  They further increase the effectiveness
of the nonlocal prior using image patches retrieved from an external
dataset.  The work of Shih et al. \cite{shih2015reflection} takes
advantage of ghosting, the phenomenon of multiple reflections caused
by thicker glass, and decomposes the input image based on Gaussian
mixture model (GMM).  To deal with reflections from eyeglasses in
frontal face image, Sandhan and Choi \cite{sandhan2017anti} exploit
the bilateral symmetry of human face and use a mirrored input image as
another input of varied reflections.

For more detailed review on the existing techniques for reflection
removal, we refer readers to two excellent surveys
\cite{artusi2011survey} and \cite{wan2017benchmarking}.

\section{Preparation of Training Data}

The proposed technique recognizes and separates reflections from the
input image using an end-to-end mapping trained by image pairs with
and without reflections.  The effectiveness of our technique, or any
machine learning approaches, greatly relies on the availability of a
representative and sufficiently large set of training data.  In this
section, we discuss the methods for collecting and preparing the
training images for our technique.

To help the proposed technique identify the patterns of reflections in
real-world scenarios, ideally, the training algorithm should only use
real photographs as the training data.  Obtaining an image with real
reflections is not difficult; we can capture such a mixture image $I$,
as in Eq.~\eqref{eq:itr}, by placing a piece of reflective glass of
transmittance $\alpha$ in front of the camera.  The corresponding
clean image $T$ of the same scene is also attainable using the same
camera setup but without the glass.  However, training images
collected using this scheme have several non-negligible drawbacks and
limitations.  First, it is almost impossible to get a pair of images
that are perfectly aligned.  Even with a tripod that stabilizes the
camera, the motions of objects within the scene can still cause
misalignment between two images captured consecutively.

Furthermore, due to the effects of refraction, the glass shielding the
scene shifts the path of light transmitting through the glass and can
also lead to the alignment problem.  By the reflection formation model
in Eq.~\eqref{eq:itr}, these differences introduced by the
misalignment between the mixture image $I$ and its reflection-free
counterpart $T$ can be seen as a part of the noise term $n$, where
\begin{equation}
  \beta R + n = I - \alpha T.
\end{equation}
Since the noise introduced by misalignment has similar characteristics
as a natural image, it is difficult to accurately distinguish the
noise $n$ from the reflection image $\beta R$.  As a result, a
training algorithm could erroneously attribute part of the noise $n$
as the effects of the reflection $\beta R$, interfering the learning
of the true reflections.  Similarly, regional illumination changes
between a pair of images can lead to the increase of structural noise
in training data as well.  Although it is possible to reduce these
adverse effects by carefully shooting only static scene from a
stationary camera or using thinner reflective glass with small
refraction, these methods greatly limit the flexibility and
practicality of collecting real images as training data.

Due to the unavoidable limitations discussed above and the prohibitive
cost of building a large enough training set of real images, we use
synthetic images constructed from images collected online for training
instead.  The main idea of the synthesis process follows the physical
reflection formation model in Eq.~\eqref{eq:itr}, which interprets a
reflection-interfered image $I$ as the linear combination of two
reflection-free natural images $T, R$.  The formation model, however,
cannot be applied directly to most of the JPEG-compressed images
available online.  The reason is that, to take advantage of human's
non-linear light sensitivity, JPEG images have to be gamma corrected
before being stored on camera, hence their pixel values are not linear
to the light intensities captured by image sensor.  Consequently, the
direct summation of two gamma-corrected images does not conform the
physics of light superposition as required by Eq.~\eqref{eq:itr},
resulting unrealistic reflection-interfered image.  To correct this
problem, we can either only use raw image or apply inverse gamma
correction on the collected JPEG images, as follows
\begin{equation}
  X = (X')^{1/\gamma},
\end{equation}
where $X'$ is a gamma corrected image and $X$ is the corresponding
light intensity image.  The gamma correction coefficient $\gamma$ for
each color channel is often available in exchangeable image file
format (EXIF) segment attached in each JPEG image.  In the following
discussion, we still use the linear formula as in Eq.~\eqref{eq:itr}
and assume that all the pixel values are restored to the raw light
intensity readings from image sensor.

To accurately simulate the formation of reflection-interfered image,
we also consider the blur effect in the reflections.  In most real
images, the focal planes of the camera are on the objects behind the
glass rather than the reflected objects, since what behind the
reflective glass are normally the objects of interest.  As a result,
reflections are often blurry due to the defocus effect
\cite{li2014single, fan2017generic}.  To simulate this effect in the
synthetic image, we blur the reflection image $R$ with a Gaussian
kernel $G$ of random variance before superimposing it into the
synthetic image $I_B$, as follows,
\begin{equation}
  I_B= \alpha T + \beta R * G
  \label{eq:ib}
\end{equation}

\begin{figure}
  \centering
  \includegraphics[width=0.7\linewidth]{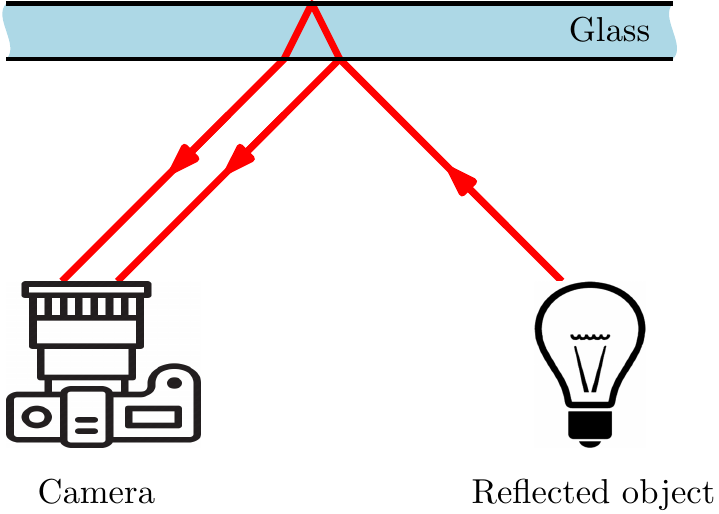}
  \caption{The formation of double reflection.}
  \label{fig:2ref}
\end{figure}

In addition to the reflection blurring, we also consider the double
reflection effect in synthesized image.  Double reflection effect is
formed due to the reflections from the two surfaces of the reflective
glass as shown in Fig.~\ref{fig:2ref}.  The offset between the two
reflection images are decided by the relative position and angle of
the camera and the reflective glass.  Suppose the transmittance of the
glass is $\alpha$ and its reflectivity is $1-\alpha$, then the
strengths of the double reflections are approximately $
1-\sqrt{\alpha}$ and $\sqrt{\alpha}-\alpha$, respectively.  This reflection effect
can be simulated by convoluting the reflection image $R$ with a random
kernel $K$ with two pulses of amplitude $1-\sqrt{\alpha}$ and
$\sqrt{\alpha}-\alpha$.

\begin{figure}
  \begin{center}
    \includegraphics[width=3.2 in]{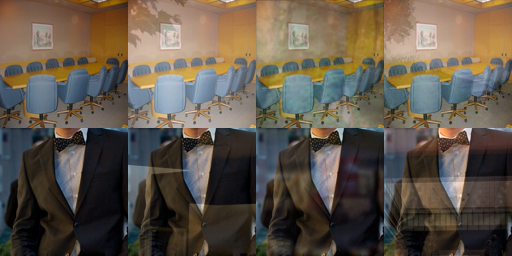}
  \end{center}
  \caption{Samples of synthesized images.  From left to right:
    Transmission images, synthetic images with sharp reflections,
    synthetic images with blurry reflections, synthetic images with
    double reflections}
  \label{fig:data_example}
\end{figure}

Combining the blurring and double reflection effects together, we
arrive at a generic formula for synthesizing reflection-interfered
image $I$.
\begin{equation}
  I = \alpha T + \beta R * G * K.
  \label{eq:i}
\end{equation}
Some samples of synthesized images are shown in
\figurename~\ref{fig:data_example}.

\section{Proposed Method}

\subsection{Network Architecture}

Similar to many deep learning based image restoration techniques, the
proposed reflection removal method adopts the basic architecture of a
convolutional encoder-decoder network \cite{mao2016image}.  The
ultimate goal of our method is to find an optimal end-to-end mapping
$T'=F(I)$ from a reflection-interfered image $I=T'+R'$ to its
transmission layer $T'$, where $R'$ is the reflection layer of $I$.
The transmission layer $T'=\alpha T$ is a glass-free image $T$ of the
targeted scene attenuated by $\alpha$, the transmittance of the
reflective glass in image $I$, as in the training data synthesis
formula Eq.~\ref{eq:i}.  Since the reflection layer $R'$ is likely
weak and smooth in comparison with $T'$ \cite{li2014single,
  arvanitopoulos2017single}, $T'$ should be similar to $I$ in pixel
values.  Therefore, it is easier to optimize mapping $T'=F(I)$ than to
optimize the mapping from $I$ to $T$ directly, even if transmittance
$\alpha$ is known to the network.  Once the solution to the
transmission layer $T'$ is given, it is trivial to restore a realistic
reflection free image $T$ from $T'$.

Another option is to train a residual mapping from image $I$ to its
reflection layer $R'=I-F(I)$.  Since $R'$ is relatively flat, the
optimization of the residual mapping should be very effective.
Residual learning has set the state of the art for many different
image restoration problems \cite{ledig2016photo, kim2016accurate,
  he2016deep}.  However, none of the existing residual learning
networks suits the characteristics of the reflection removal problem.
In most residual learning techniques, the residual is the missing
detail to be recovered and added back to the input image, while in our
case, the residual is unwanted reflection that should be subtracted
from the input.  Furthermore, residual learning tends to emphasize on
the fidelity of the recovered residual as the loss function is
normally applied on the residual.  But for the reflection removal
problem, as long as the recovered transmission layer has reduced
interference and looks natural, the restoration quality of the
reflection layer is irrelevant.  Due to these limitations, it is
difficult to get satisfactory results for reflection removal by using
residual learning directly.  Thus, we place a residual learning based
reflection recovery sub-network at the middle of our end-to-end
mapping $T'=F(I)$ network, in order to exploit the efficiency of
residual learning without affecting the output quality.

\begin{figure*}
  \begin{center}
    \includegraphics[width=6.5in]{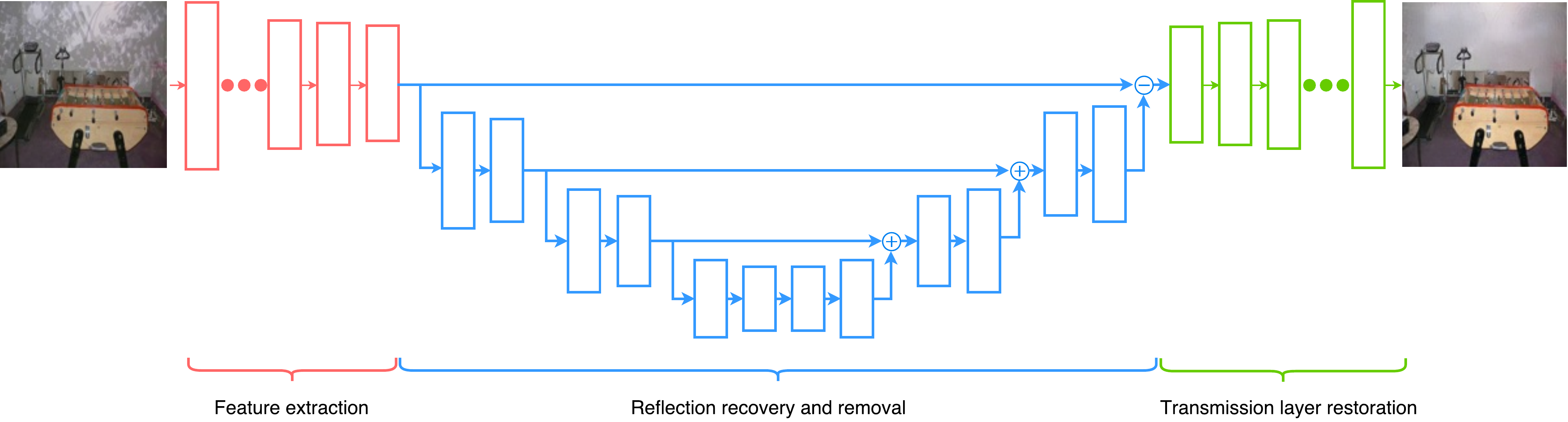}
  \end{center}
  \caption{Architecture of the used convolutional auto-encoder with symmetric shortcut connection.}
  \label{fig:network}
\end{figure*}

\begin{figure}
  \centering
  \begin{subfigure}{0.32\linewidth}
    \centering
    \includegraphics[width=1.0in]{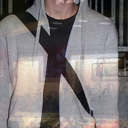}
    \caption{Synthetic input\\ \ }
    \label{fig:a}
  \end{subfigure}
  \begin{subfigure}{0.32\linewidth}
    \centering
    \includegraphics[width=1.0in]{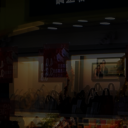}
    \caption{Recovered reflection image}
    \label{fig:b}
  \end{subfigure}
  \begin{subfigure}{0.32\linewidth}
    \centering
    \includegraphics[width=1.0in]{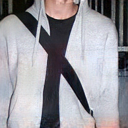}
    \caption{Recovered transmission image}
    \label{fig:c}
  \end{subfigure}
  \begin{subfigure}{0.32\linewidth}
    \centering
    \includegraphics[width=1.0in]{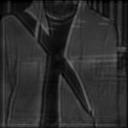}
    \caption{$\mathit{conv}_2$}
    \label{fig:d}
  \end{subfigure}
  \begin{subfigure}{0.32\linewidth}
    \centering
    \includegraphics[width=1.0in]{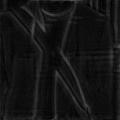}
    \caption{$\mathit{conv}_4$}
    \label{fig:e}
  \end{subfigure}
  \begin{subfigure}{0.32\linewidth}
    \centering
    \includegraphics[width=1.0in]{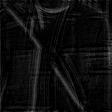}
    \caption{$\mathit{conv}_6$}
    \label{fig:f}
  \end{subfigure}
  \begin{subfigure}{0.32\linewidth}
    \centering
    \includegraphics[width=1.0in]{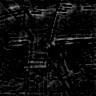}
    \caption{$\mathit{deconv}_2$}
    \label{fig:g}
  \end{subfigure}
  \begin{subfigure}{0.32\linewidth}
    \centering
    \includegraphics[width=1.0in]{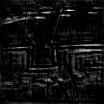}
    \caption{$\mathit{deconv}_4$}
    \label{fig:h}
  \end{subfigure}
  \begin{subfigure}{0.32\linewidth}
    \centering
    \includegraphics[width=1.0in]{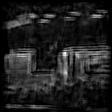}
    \caption{$\mathit{deconv}_6$}
    \label{fig:i}
  \end{subfigure}
  \begin{subfigure}{0.32\linewidth}
    \centering
    \includegraphics[width=1.0in]{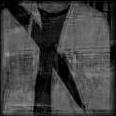}
    \caption{$\mathit{deconv}_7$}
    \label{fig:j}
  \end{subfigure}
  \begin{subfigure}{0.32\linewidth}
    \centering
    \includegraphics[width=1.0in]{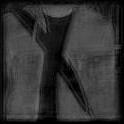}
    \caption{$\mathit{deconv}_9$}
    \label{fig:k}
  \end{subfigure}
  \begin{subfigure}{0.32\linewidth}
    \centering
    \includegraphics[width=1.0in]{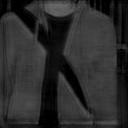}
    \caption{$\mathit{deconv}_{11}$}
    \label{fig:l}
  \end{subfigure}
  \caption{Sample feature maps at different stages of the network.
    These feature maps show that the network is working as intended in
    each stage.}
  \label{fig:feature maps}
\end{figure}

\begin{figure}
  \begin{center}
    \includegraphics[width=3.0in]{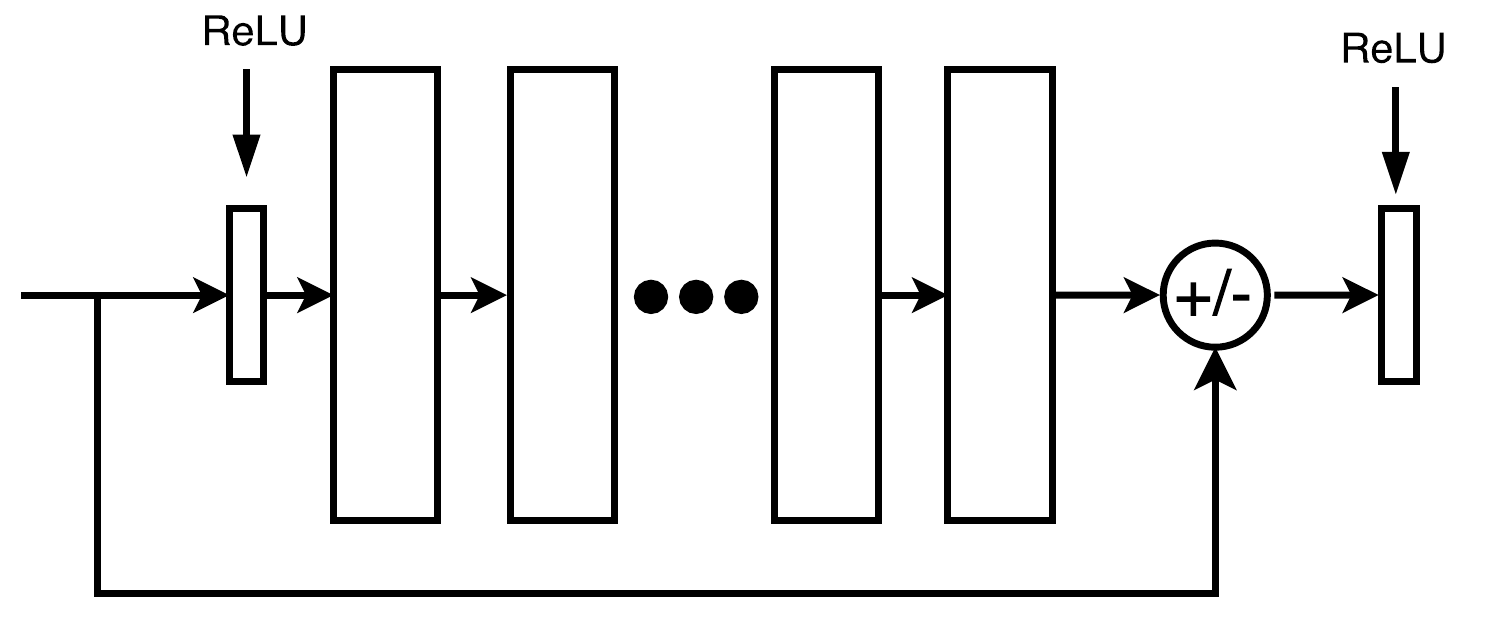}
  \end{center}
  \caption{The topology of a single shortcut connection.}
  \label{fig:shortcut}
\end{figure}

The proposed network consists of 12 convolutional layers and 12
deconvolutional with one rectified linear unit (ReLU) following each
of the layers.  The convolutional layers are designed to extract and
condense features from the input, while deconvolutional layers rebuild
the details of reflection-free image from feature abstractions.
Overall, the architecture of proposed encoder-decoder network can be
divided into three stages:
\begin{enumerate}
\item \textbf{Feature extraction}. The 6 convolutional layers in this
  stage extract features for both transmission and reflection layers
  as illustrated in \figurename~\ref{fig:network}.  The outputs of the
  convolutional layers are shown in the second row of
  \figurename~\ref{fig:feature maps}.
\item \textbf{Reflection recovery and removal}.  In this stage, the
  first 6 convolutional layers and following 6 deconvolutional layers
  are set to learn and recover reflection.  Additionally, to preserve
  the details of the reflection layer better, two skip connections are
  added in the second stage to inherit the features learned from
  previous convolutional layers \cite{mao2016image}.  The topology of
  skip connection is illustrated in \figurename~\ref{fig:shortcut}.
  At the end of this stage, the recovered reflection is removed before
  $\mathit{deconv}_7$, the seventh deconvolutional layer, by using an
  element-wise subtraction \cite{yang2016joint} followed by a ReLU
  activation $\max (0, \mathit{conv}_6 \!-\!  \mathit{deconv}_6)$.  As
  shown in the third row of \figurename~\ref{fig:feature maps}, this
  stage removes the transmission layer gradually from the input and
  preserves only the features of the reflections.
\item \textbf{Transmission layer restoration}.  The reflections might
  not be removed completely after previous stage by simply subtracting
  the estimated reflection, as shown in the last row of
  \figurename~\ref{fig:feature maps}.  Thus, this stage tries to
  restore a visually pleasing transmission image from the reflection
  subtracted image.  To achieve this goal, 6 deconvolutional layers
  are used to recover transmission layer from the features of the
  targeted scene.
\end{enumerate}

For image classification tasks, pooling layers are necessary as it
extracts main abstract features that are crucial for final decision
\cite{he2016deep}.  However, as the redundant information increases
the difficulty for deconvolutional layers to recover the image
\cite{mao2016image}, pooling layers are omitted in our reflection
removal network.  Another important factor that affects the
performance of our encoder-decoder network is the size of the
convolutional kernel.  To make the network learn the semantic context
of an image, we employ relatively large kernels ($5\times5$).  But if
the input image contains double reflections of large disparity, we
find a larger kernel ($9 \times 9$) for the first convolutional layers
($\mathit{conv}_1$ and $\mathit{conv}_2$) and the last deconvolutional
layers ($\mathit{deconv}_{11}$ and $\mathit{deconv}_{12}$) is
necessary to achieve the best performance.

\subsection{Loss functions}

In many neural network based image restoration techniques, such as
denoising \cite{xie2012image}, debluring \cite{xu2014deep} and super
resolution \cite{kim2016accurate}, the networks are commonly optimized
using mean square error (MSE) between the output and the ground truth
as the loss function,
\begin{align}
  L_{\ell_2}=\|F(I)-\alpha T\|_2^2
  \label{eq:l2}
\end{align}
However, a model optimized using only $\ell_2$-norm loss function
often fails to preserve high-frequency contents.  In the case of
reflection removal, both the reflection and transmission layers are
natural images with different characteristics.  To get the best
restoration result, the network should learn the perceptual properties
of the transmission layer.  Inspired by \cite{gatys2015texture,
  johnson2016perceptual}, we employ a loss function that is closer to
high-level feature abstractions.  Based on Ledig
et~al. \cite{ledig2016photo}, the VGG loss is calculated as the
$\ell_2$-norm of the difference between the layer representations of
the restored transmission $T'=F(I)$ and the real transmission image
$\alpha T$ on the pre-trained 19 layers VGG network proposed by
Simonyan and Zisserman \cite{simonyan2014very}:
\begin{align}
  L_{\text{VGG}}=\frac{1}{W_{i}H_{i}} \sum_{i=1}^M
  \|\phi_{i}(\alpha T)-\phi_{i}(F(I))\|_2^2
\end{align}
where $\phi_{i}$ is the feature maps obtained by the $i$-th
convolution layer (after activation) within the VGG19 network; $M$ is
the number of convolution layers used; and $W_{i}$ and $H_{i}$ are the
dimension of $i$-th feature map.  In our model, the feature maps of
the first 5 convolution layers are used ($M=5$) to build the
perceptual loss.  The final loss for training is calculated as:
\begin{align}
  L=L_{\ell_2}+\lambda L_{\text{VGG}},
  \label{eq:l2+vgg}
\end{align}
where $\lambda$ is a parameter for balancing the contributions from
the $\ell_2$-norm loss and the VGG loss.

\begin{figure}
  \begin{center}
    \includegraphics[width=\linewidth]{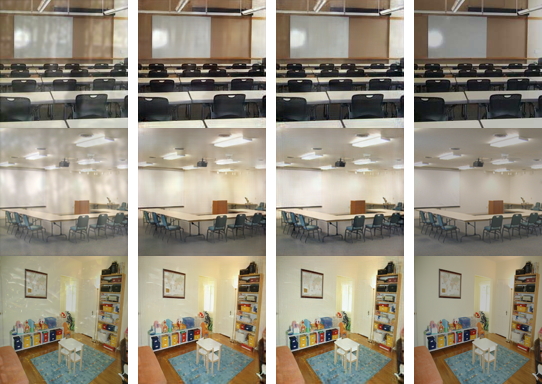}
  \end{center}
  \caption{From left to right: input reflection-interfered images,
    network optimized for $\ell_2$-norm loss, network optimized for
    both $\ell_2$-norm and VGG losses, the ground-truth transmission
    images.}
  \label{fig:comparison_vgg_l2}
\end{figure}

\figurename~\ref{fig:comparison_vgg_l2} presents some of the results
from two reflection removal models trained with loss functions in
Eqs.~\eqref{eq:l2} and \eqref{eq:l2+vgg} respectively.  As shown in
the figure, the output images (3rd column) from the model trained with
the mixed loss function as in Eq.~\eqref{eq:l2+vgg} is much cleaner
and closer to the ground-truths (4th column) than those output images
(2nd column) from the model trained solely with $\ell_2$-norm loss as
in Eq.~\eqref{eq:l2}.

\section{Experiments}

In this section, we first discuss how the training sets are formed and
how the parameters are set for training the network.  Then, we
evaluate the proposed method using synthetic and real images and
compare our results with several state-of-art methods.

\begin{figure*}[t]
  \centering
  \captionsetup[subfigure]{labelformat=empty}
  \begin{subfigure}{0.19\linewidth}
    \includegraphics[width=1.3in]{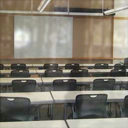}
  \end{subfigure}
  \begin{subfigure}{0.19\linewidth}
    \includegraphics[width=1.3in]{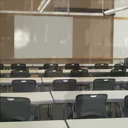}
  \end{subfigure}
  \begin{subfigure}{0.19\linewidth}
    \includegraphics[width=1.3in]{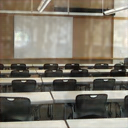}
  \end{subfigure}
  \begin{subfigure}{0.19\linewidth}
    \includegraphics[width=1.3in]{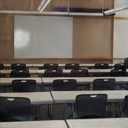}
  \end{subfigure}
  \begin{subfigure}{0.19\linewidth}
    \includegraphics[width=1.3in]{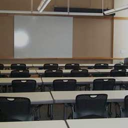}
  \end{subfigure}

  \begin{subfigure}{0.19\linewidth}
    \includegraphics[width=1.3in]{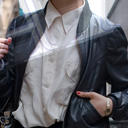}
    \caption{Synthetic input}
  \end{subfigure}
  \begin{subfigure}{0.19\linewidth}
    \includegraphics[width=1.3in]{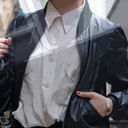}
    \caption{\cite{arvanitopoulos2017single}}
  \end{subfigure}
  \begin{subfigure}{0.19\linewidth}
    \includegraphics[width=1.3in]{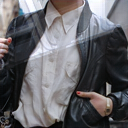}
    \caption{\cite{fan2017generic}}
  \end{subfigure}
  \begin{subfigure}{0.19\linewidth}
    \includegraphics[width=1.3in]{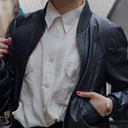}
    \caption{Ours}
  \end{subfigure}
  \begin{subfigure}{0.19\linewidth}
    \includegraphics[width=1.3in]{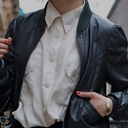}
    \caption{Ground truth}
  \end{subfigure}

  \caption{Comparison of reflection removal algorithms using synthetic images.}

  \label{synthetic image compare}
\end{figure*}

\subsection{Data Preparation}

To simulate the scenarios where reflections interfere the formation of
images, 2303 images from the indoor scene recognition dataset
\cite{quattoni2009recognizing} and 2622 street snap images
\cite{jokinen2017hellooks} are collected online.  We choose the images
of natural landscapes and images taken inside a mall as the
reflections.  Leaves could create sparse shadow on the window
interfering the transmission image, and the lights from a shop sign
create strong and sharp reflection, which are also extremely common in
real life reflection-interfered images.  To ensure the size of
training dataset and avoid over-fitting, each transmission image is
synthesized with 18 randomly chosen reflection images using
Eq.~\eqref{eq:i}.  To simulate the different blurriness of the
reflections, the variance of the Gaussian blur kernel is selected
randomly from 1 to 5.  The transmittance $\alpha$ is also a random
number between 0.75 to 0.8 for each synthetic image.  Before
generating a synthetic image, the transmission layer is resized to
$128 \times 128$, whereas the reflection layer is randomly cropped
from a large reflection image and then resized to $128 \times 128$.
The reason for this step is that the reflected objects are normally
far away from the glass, as a result, larger objects in the reflection
scene appear relatively smaller in the reflection-interfered image.
Finally, all the synthesized images are split into a training set of
66540 images and a testing set of 22110 images.


\subsection{Network Training}

The network is trained in an end-to-end manner as $T'=F(I)$ and
parametrized by $\theta_F$ which donates weights and biases of the
network. It is aimed to solve the following objective
function:
\begin{align}
  \theta_F=\arg\min_{\theta_F}\frac{1}{N}\sum_{n=1}^N L(F(I_n),\alpha T_n)
\end{align}
where N is the size of training set, $L$ is the combined loss
functions defined in Eq.~\ref{eq:l2+vgg}.  We set 64 filters for all
the convolutional and deconvolutional layers and the VGG loss weight
$\lambda$ is set to $0.001$.  The network is optimized by Adam
optimizer \cite{kingma2014adam} with a learning rate of $10^{-4}$ and
$\beta_1 = 0.9$, batch size is set to 64 to accelerate the training
process.  The network is trained for 150 epochs. All the experiments
are done on a NVIDIA TITAN X GPU and implementation is based on
Tensorflow \cite{tensorflow2015-whitepaper}.

\begin{figure*}[ht]
  \centering
  \captionsetup[subfigure]{labelformat=empty}
  \begin{subfigure}{0.12\linewidth}
    \includegraphics[width=0.83in]{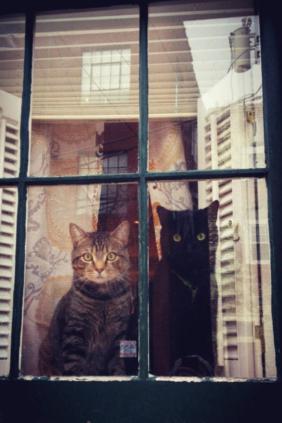}
  \end{subfigure}
  \begin{subfigure}{0.12\linewidth}
    \includegraphics[width=0.83in]{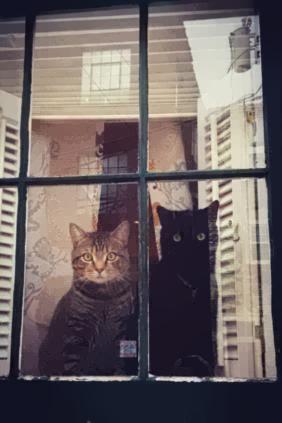}
  \end{subfigure}
  \begin{subfigure}{0.12\linewidth}
    \includegraphics[width=0.83in]{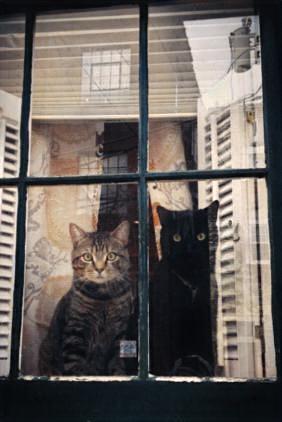}
  \end{subfigure}
  \begin{subfigure}{0.12\linewidth}
    \includegraphics[width=0.83in]{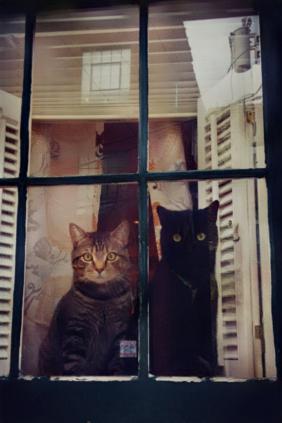}
  \end{subfigure}
  \begin{subfigure}{0.12\linewidth}
    \includegraphics[width=0.83in,height=1.25in]{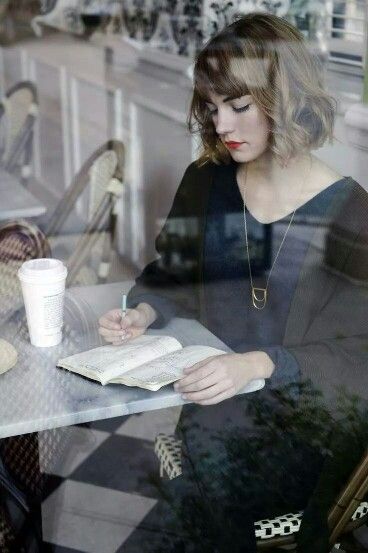}
  \end{subfigure}
  \begin{subfigure}{0.12\linewidth}
    \includegraphics[width=0.83in,height=1.25in]{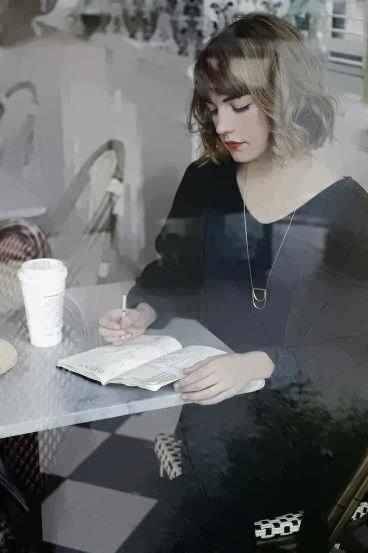}
  \end{subfigure}
  \begin{subfigure}{0.12\linewidth}
    \includegraphics[width=0.83in,height=1.25in]{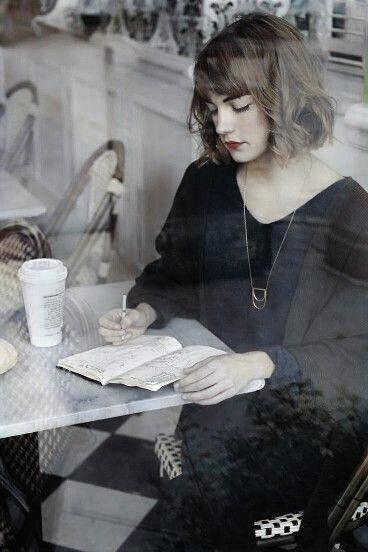}
  \end{subfigure}
  \begin{subfigure}{0.12\linewidth}
    \includegraphics[width=0.83in,height=1.25in]{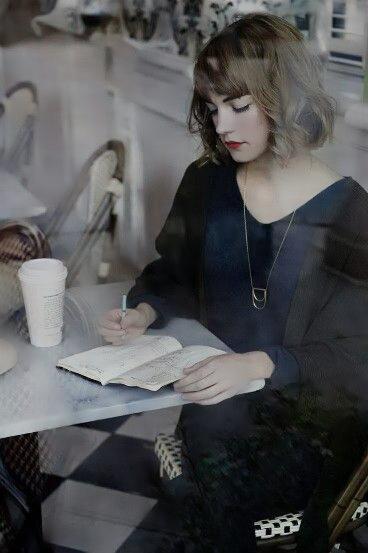}
  \end{subfigure}
  \begin{subfigure}{0.12\linewidth}
    \includegraphics[width=0.83in,height=1.05in]{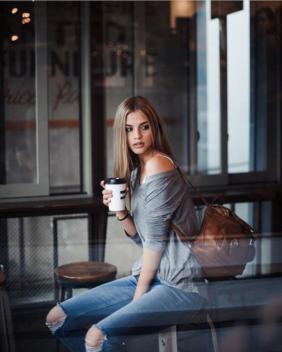}
    \caption{Input}
  \end{subfigure}
  \begin{subfigure}{0.12\linewidth}
    \includegraphics[width=0.83in,height=1.05in]{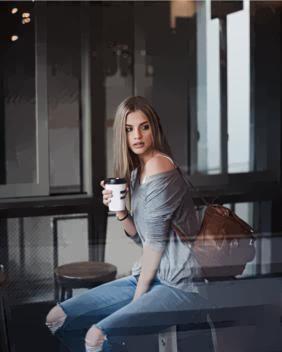}
    \caption{\cite{arvanitopoulos2017single}}
  \end{subfigure}
  \begin{subfigure}{0.12\linewidth}
    \includegraphics[width=0.83in,height=1.05in]{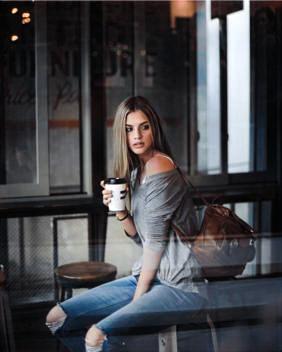}
    \caption{\cite{fan2017generic}}
  \end{subfigure}
  \begin{subfigure}{0.12\linewidth}
    \includegraphics[width=0.83in,height=1.05in]{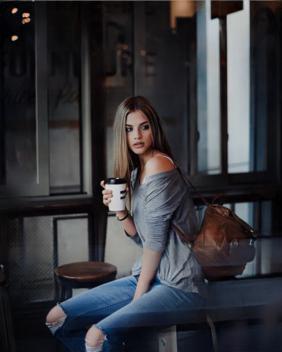}
    \caption{Ours}
  \end{subfigure}
  \begin{subfigure}{0.12\linewidth}
    \includegraphics[width=0.83in,height=1.05in•]{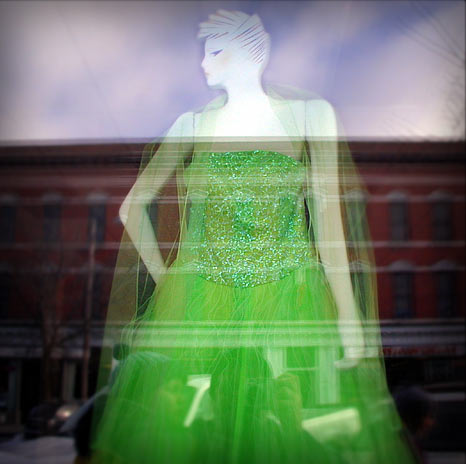}
    \caption{Input}
  \end{subfigure}
  \begin{subfigure}{0.12\linewidth}
    \includegraphics[width=0.83in,height=1.05in]{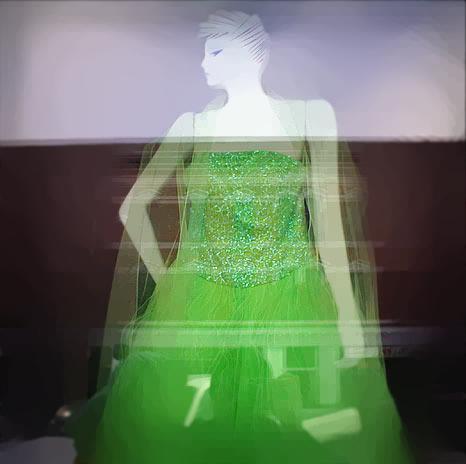}
    \caption{\cite{arvanitopoulos2017single}}
  \end{subfigure}
  \begin{subfigure}{0.12\linewidth}
    \includegraphics[width=0.83in,height=1.05in]{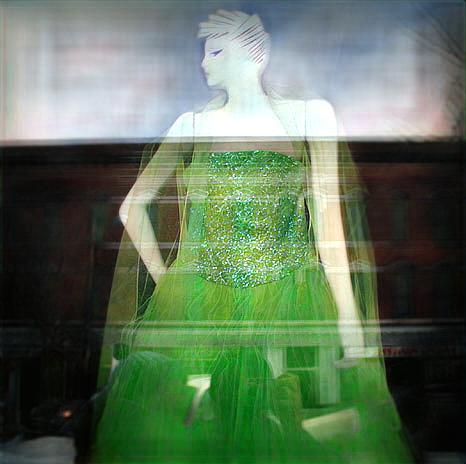}
    \caption{\cite{fan2017generic}}
  \end{subfigure}
  \begin{subfigure}{0.12\linewidth}
    \includegraphics[width=0.83in,height=1.05in]{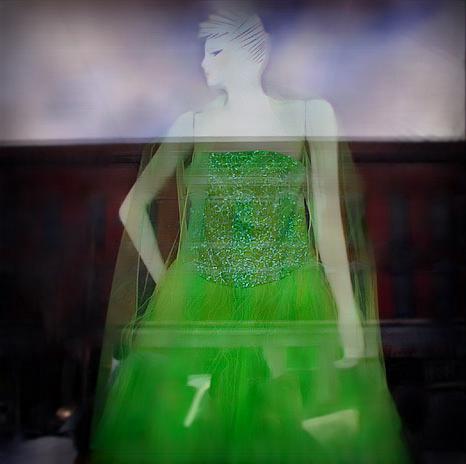}
    \caption{Ours}
  \end{subfigure}

  \begin{subfigure}{0.99\linewidth}
    \rule{17.5cm}{0.05cm}
  \end{subfigure}
  \begin{subfigure}{0.24\linewidth}
    \includegraphics[width=1.66in,height=1.1in]{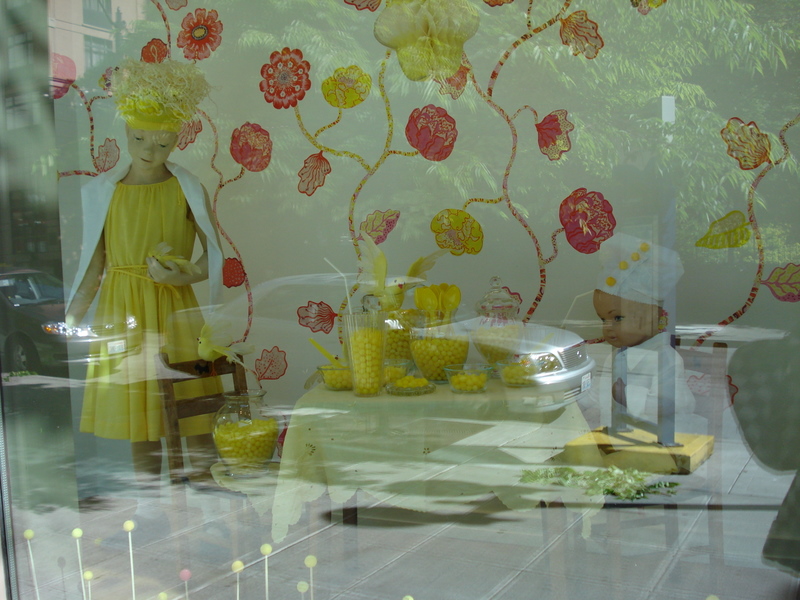}
  \end{subfigure}
  \begin{subfigure}{0.24\linewidth}
    \includegraphics[width=1.66in,height=1.1in]{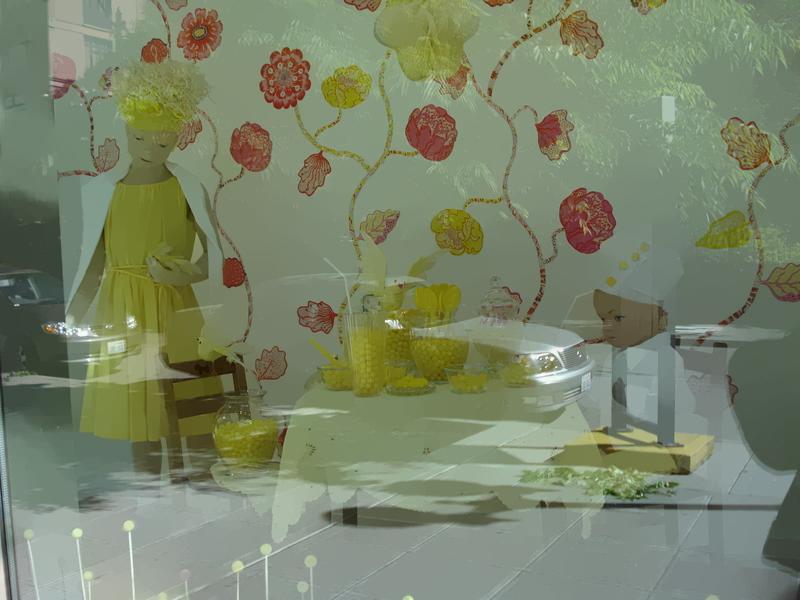}
  \end{subfigure}
  \begin{subfigure}{0.24\linewidth}
    \includegraphics[width=1.66in,height=1.1in]{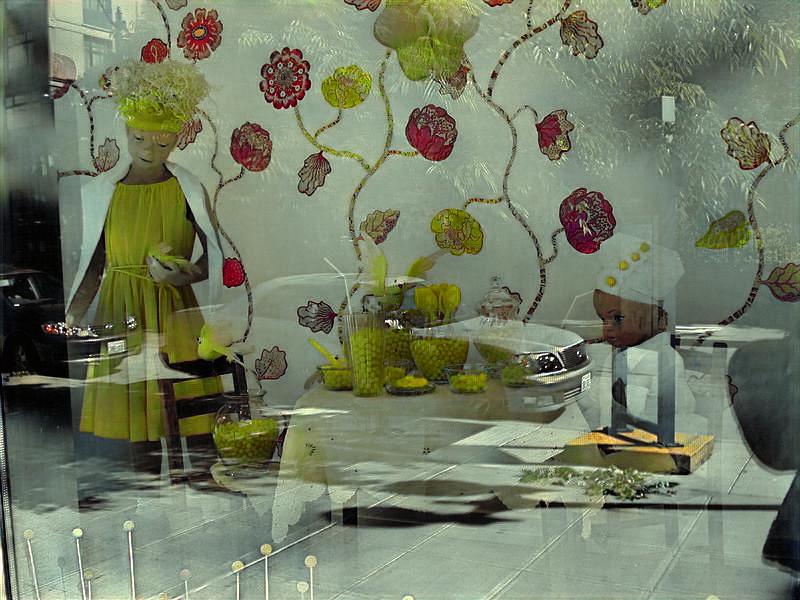}
  \end{subfigure}
  \begin{subfigure}{0.24\linewidth}
    \includegraphics[width=1.66in,height=1.1in]{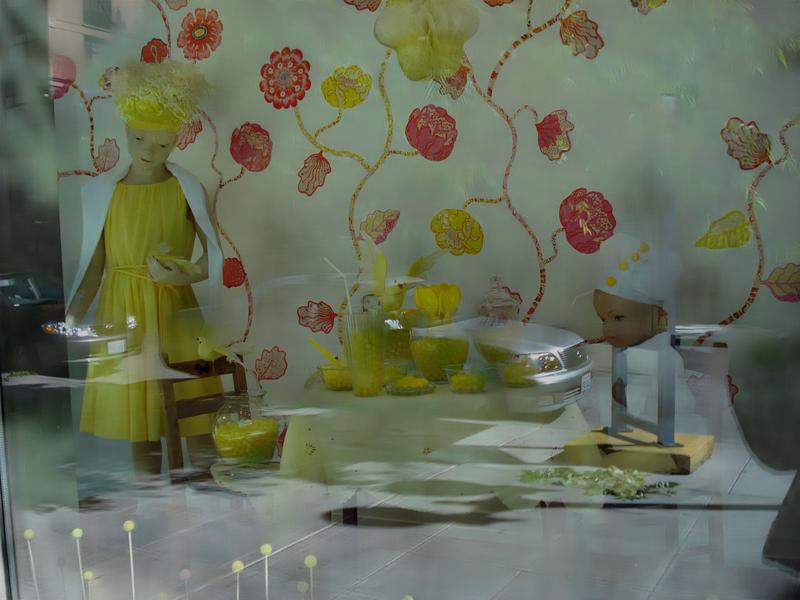}
  \end{subfigure}

  \begin{subfigure}{0.24\linewidth}
    \includegraphics[width=1.66in,height=1.05in]{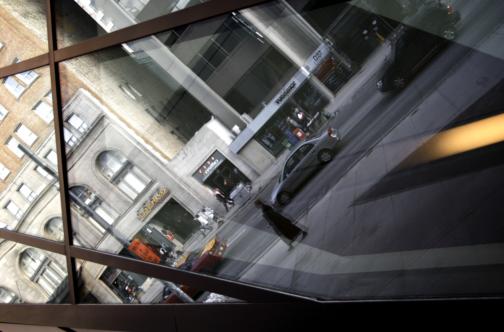}
  \end{subfigure}
  \begin{subfigure}{0.24\linewidth}
    \includegraphics[width=1.66in,height=1.05in]{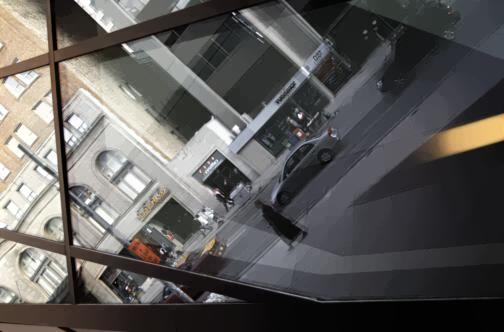}
  \end{subfigure}
  \begin{subfigure}{0.24\linewidth}
    \includegraphics[width=1.66in,height=1.05in]{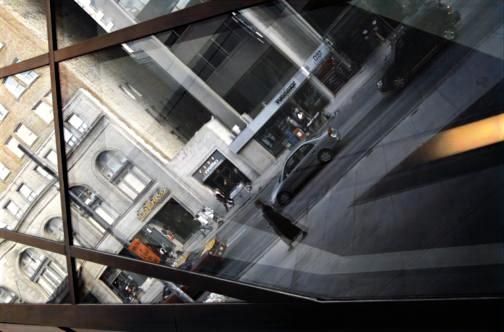}
  \end{subfigure}
  \begin{subfigure}{0.24\linewidth}
    \includegraphics[width=1.66in,height=1.05in]{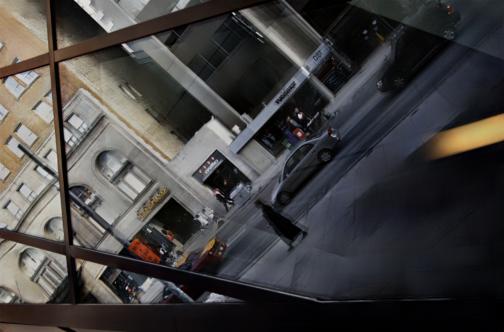}
  \end{subfigure}

  \begin{subfigure}{0.24\linewidth}
    \includegraphics[width=1.66in,height=1.05in]{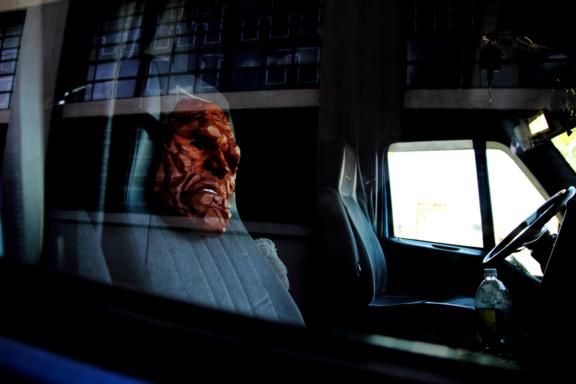}
    \caption{Input}
  \end{subfigure}
  \begin{subfigure}{0.24\linewidth}
    \includegraphics[width=1.66in,height=1.05in]{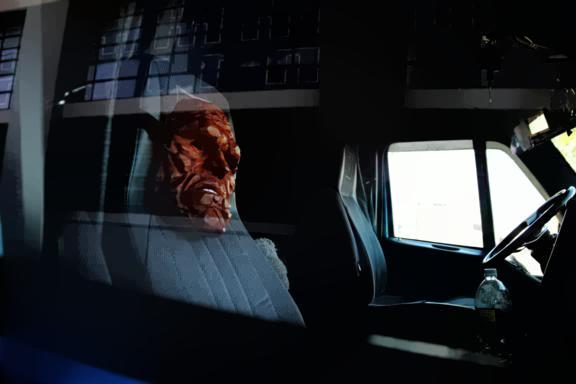}
    \caption{\cite{arvanitopoulos2017single}}
  \end{subfigure}
  \begin{subfigure}{0.24\linewidth}
    \includegraphics[width=1.66in,height=1.05in]{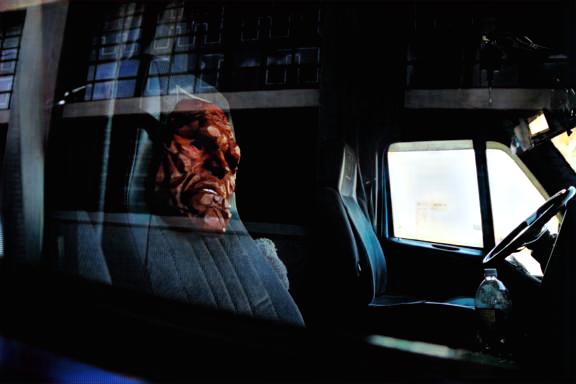}
    \caption{\cite{fan2017generic}}
  \end{subfigure}
  \begin{subfigure}{0.24\linewidth}
    \includegraphics[width=1.66in,height=1.05in]{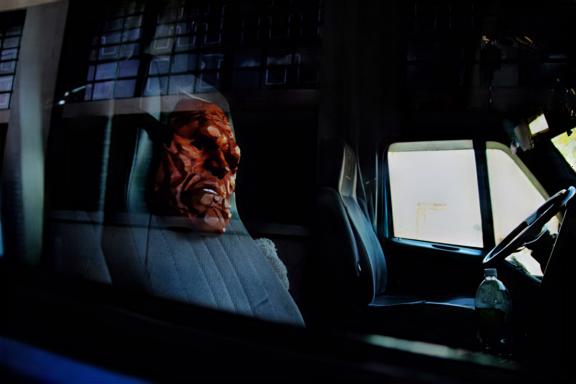}
    \caption{Ours}
  \end{subfigure}

  \caption{Comparison of reflection removal algorithms using real
    images.}

  \label{Real-life image compare}
\end{figure*}

\subsection{Evaluation}

\figurename~\ref{synthetic image compare} shows some results of the
proposed method in comparison with the results of two state-of-the-art
reflection removal techniques \cite{arvanitopoulos2017single} and
\cite{fan2017generic} using synthetic data.  It is expected that the
proposed method significantly outperforms the compared techniques in
this case, as our network is trained with similar images generated by
the data synthesizer.  For real-life images collected by us or
provided by the authors of \cite{arvanitopoulos2017single}, the
results of proposed method are still the best among the tested
techniques, as shown in \figurename~\ref{Real-life image compare}.
The technique in \cite{fan2017generic} is based on the assumption that
the reflection is smooth.  However, if this assumption is not true, as
exemplified in the sample images, \cite{fan2017generic} could even
enhance the reflections, making the results worse than the inputs.
Our method does not rely on such an assumption; it works well
regardless the smoothness of the reflections.  The problem of
\cite{arvanitopoulos2017single} is the severe loss of details in its
output, resulting unnatural looking image.  In the cases where the
reflections are much stronger than the transmission layer, none of the
tested algorithms can yield satisfactory results.

\begin{table}
  \centering
  \begin{tabular}{|c|c|c|c|}
    \hline
    & \cite{arvanitopoulos2017single} & \cite{fan2017generic} & Ours \\
    \hline\hline
    Synthetic Images & 19.72 & 19.82 & \textbf{29.08} \\
    \hline
    Benchmark Set \cite{wan2017benchmarking} & 16.85 & 18.29 & \textbf{18.70} \\
    \hline
  \end{tabular}
  \caption{PSNR results of tested techniques using synthetic images
    and a benchmark dataset.}
  \label{tab:psnr}
\end{table}

Reported in Table~\ref{tab:psnr} are the PSNR results of the tested
techniques.  In addition to the synthetic images, we tested a
synthetic benchmark set provided by the authors of
\cite{wan2017benchmarking}.  Our method achieves the highest average
PSNR in these tests.  The running times of the proposed method are
comparable to other deep learning based image restoration techniques.
The proposed method takes around 0.6 s to process a $128 \times 128$
image and 2 s to process a $512 \times 512$ image.

\section{Conclusion}

The task of removing reflection interference from a single image is a
highly ill-posed problem.  We propose a new reflection formation model
taking into the consideration of physics of digital camera imaging,
and apply the model in a deep convolutional encoder-decoder network
based data-driven technique.  Extensive experimental results show
that, although the neural network learns only from synthetic data, the
proposed method is effective on real-world images, and it
significantly outperforms the other tested state-of-the-art
techniques.

{\small
\bibliographystyle{ieee}
\bibliography{reflection}

\begin{thebibliography}{10}\itemsep=-1pt

\bibitem{tensorflow2015-whitepaper}
M.~Abadi, A.~Agarwal, P.~Barham, E.~Brevdo, Z.~Chen, C.~Citro, G.~S. Corrado,
  A.~Davis, J.~Dean, M.~Devin, S.~Ghemawat, I.~Goodfellow, A.~Harp, G.~Irving,
  M.~Isard, Y.~Jia, R.~Jozefowicz, L.~Kaiser, M.~Kudlur, J.~Levenberg,
  D.~Man\'{e}, R.~Monga, S.~Moore, D.~Murray, C.~Olah, M.~Schuster, J.~Shlens,
  B.~Steiner, I.~Sutskever, K.~Talwar, P.~Tucker, V.~Vanhoucke, V.~Vasudevan,
  F.~Vi\'{e}gas, O.~Vinyals, P.~Warden, M.~Wattenberg, M.~Wicke, Y.~Yu, and
  X.~Zheng.
\newblock {TensorFlow}: Large-scale machine learning on heterogeneous systems.
\newblock \url{http://tensorflow.org/}, 2015.

\bibitem{agrawal2005removing}
A.~Agrawal, R.~Raskar, S.~K. Nayar, and Y.~Li.
\newblock Removing photography artifacts using gradient projection and
  flash-exposure sampling.
\newblock {\em ACM Transactions on Graphics (TOG)}, 24(3):828--835, 2005.

\bibitem{akashi2014separation}
Y.~Akashi and T.~Okatani.
\newblock Separation of reflection components by sparse non-negative matrix
  factorization.
\newblock In {\em Asian Conference on Computer Vision (ACCV)}, pages 611--625,
  2014.

\bibitem{artusi2011survey}
A.~Artusi, F.~Banterle, and D.~Chetverikov.
\newblock A survey of specularity removal methods.
\newblock {\em Computer Graphics Forum}, 30(8):2208--2230, 2011.

\bibitem{arvanitopoulos2017single}
N.~Arvanitopoulos~Darginis, R.~Achanta, and S.~S{\"u}sstrunk.
\newblock Single image reflection suppression.
\newblock In {\em IEEE Conference on Computer Vision and Pattern Recognition
  (CVPR)}, 2017.

\bibitem{chandramouli2016convnet}
P.~Chandramouli, M.~Noroozi, and P.~Favaro.
\newblock Convnet-based depth estimation, reflection separation and deblurring
  of plenoptic images.
\newblock In {\em Asian Conference on Computer Vision (ACCV)}, pages 129--144,
  2016.

\bibitem{fan2017generic}
Q.~Fan, J.~Yang, G.~Hua, B.~Chen, and D.~Wipf.
\newblock A generic deep architecture for single image reflection removal and
  image smoothing.
\newblock In {\em Proceedings of the IEEE International Conference on Computer
  Vision (ICCV)}, 2017.

\bibitem{farid1999separating}
H.~Farid and E.~H. Adelson.
\newblock Separating reflections and lighting using independent components
  analysis.
\newblock In {\em IEEE Conference on Computer Vision and Pattern Recognition
  (CVPR)}, volume~1, pages 262--267, 1999.

\bibitem{feris2004specular}
R.~Feris, R.~Raskar, K.-H. Tan, and M.~Turk.
\newblock Specular reflection reduction with multi-flash imaging.
\newblock In {\em Brazilian Symposium on Computer Graphics and Image
  Processing}, pages 316--321, 2004.

\bibitem{gai2012blind}
K.~Gai, Z.~Shi, and C.~Zhang.
\newblock Blind separation of superimposed moving images using image
  statistics.
\newblock {\em IEEE Transactions on Pattern Analysis and Machine Intelligence
  (PAMI)}, 34(1):19--32, 2012.

\bibitem{gatys2015texture}
L.~Gatys, A.~S. Ecker, and M.~Bethge.
\newblock Texture synthesis using convolutional neural networks.
\newblock In {\em Advances in Neural Information Processing Systems (NIPS)},
  pages 262--270, 2015.

\bibitem{guo2014robust}
X.~Guo, X.~Cao, and Y.~Ma.
\newblock Robust separation of reflection from multiple images.
\newblock In {\em IEEE Conference on Computer Vision and Pattern Recognition
  (CVPR)}, pages 2187--2194, 2014.

\bibitem{han2017reflection}
B.-J. Han and J.-Y. Sim.
\newblock Reflection removal using low-rank matrix completion.
\newblock In {\em IEEE Conference on Computer Vision and Pattern Recognition
  (CVPR)}, pages 5438--5446, 2017.

\bibitem{he2016deep}
K.~He, X.~Zhang, S.~Ren, and J.~Sun.
\newblock Deep residual learning for image recognition.
\newblock In {\em IEEE Conference on Computer Vision and Pattern Recognition
  (CVPR)}, pages 770--778, 2016.

\bibitem{johnson2016perceptual}
J.~Johnson, A.~Alahi, and L.~Fei-Fei.
\newblock Perceptual losses for real-time style transfer and super-resolution.
\newblock In {\em European Conference on Computer Vision (ECCV)}, pages
  694--711, 2016.

\bibitem{jokinen2017hellooks}
L.~Jokinen and K.~Sampo.
\newblock Hel looks.
\newblock \url{https://www.hel-looks.com/}.
\newblock [Online; accessed 19-Oct-2017].

\bibitem{kim2016accurate}
J.~Kim, J.~Kwon~Lee, and K.~Mu~Lee.
\newblock Accurate image super-resolution using very deep convolutional
  networks.
\newblock In {\em IEEE Conference on Computer Vision and Pattern Recognition
  (CVPR)}, pages 1646--1654, 2016.

\bibitem{kingma2014adam}
D.~Kingma and J.~Ba.
\newblock Adam: A method for stochastic optimization.
\newblock {\em arXiv preprint arXiv:1412.6980}, 2014.

\bibitem{kong2014physically}
N.~Kong, Y.-W. Tai, and J.~S. Shin.
\newblock A physically-based approach to reflection separation: from physical
  modeling to constrained optimization.
\newblock {\em IEEE Transactions on Pattern Analysis and Machine Intelligence
  (PAMI)}, 36(2):209--221, 2014.

\bibitem{ledig2016photo}
C.~Ledig, L.~Theis, F.~Husz{\'a}r, J.~Caballero, A.~Cunningham, A.~Acosta,
  A.~Aitken, A.~Tejani, J.~Totz, Z.~Wang, et~al.
\newblock Photo-realistic single image super-resolution using a generative
  adversarial network.
\newblock {\em arXiv preprint arXiv:1609.04802}, 2016.

\bibitem{levin2007user}
A.~Levin and Y.~Weiss.
\newblock User assisted separation of reflections from a single image using a
  sparsity prior.
\newblock {\em IEEE Transactions on Pattern Analysis and Machine Intelligence
  (PAMI)}, 29(9):1647--1654, 2007.

\bibitem{levin2004separating}
A.~Levin, A.~Zomet, and Y.~Weiss.
\newblock Separating reflections from a single image using local features.
\newblock In {\em IEEE Conference on Computer Vision and Pattern Recognition
  (CVPR)}, volume~1, pages 306--313, 2004.

\bibitem{li2013exploiting}
Y.~Li and M.~S. Brown.
\newblock Exploiting reflection change for automatic reflection removal.
\newblock In {\em IEEE International Conference on Computer Vision (ICCV)},
  pages 2432--2439, 2013.

\bibitem{li2014single}
Y.~Li and M.~S. Brown.
\newblock Single image layer separation using relative smoothness.
\newblock In {\em IEEE Conference on Computer Vision and Pattern Recognition
  (CVPR)}, pages 2752--2759, 2014.

\bibitem{mao2016image}
X.-J. Mao, C.~Shen, and Y.-B. Yang.
\newblock Image restoration using convolutional auto-encoders with symmetric
  skip connections.
\newblock {\em arXiv preprint arXiv:1606.08921}, 2016.

\bibitem{ni2017reflection}
Y.~Ni, J.~Chen, and L.-P. Chau.
\newblock Reflection removal based on single light field capture.
\newblock In {\em IEEE International Symposium on Circuits and Systems
  (ISCAS)}, pages 1--4, 2017.

\bibitem{ohnishi1996separating}
N.~Ohnishi, K.~Kumaki, T.~Yamamura, and T.~Tanaka.
\newblock Separating real and virtual objects from their overlapping images.
\newblock In {\em European Conference on Computer Vision (ECCV)}, pages
  636--646, 1996.

\bibitem{quattoni2009recognizing}
A.~Quattoni and A.~Torralba.
\newblock Recognizing indoor scenes.
\newblock In {\em IEEE Conference on Computer Vision and Pattern Recognition
  (CVPR)}, pages 413--420, 2009.

\bibitem{sandhan2017anti}
T.~Sandhan and J.~Y. Choi.
\newblock Anti-glare: Tightly constrained optimization for eyeglass reflection
  removal.
\newblock In {\em IEEE Conference on Computer Vision and Pattern Recognition
  (CVPR)}, pages 1241--1250, 2017.

\bibitem{sarel2004separating}
B.~Sarel and M.~Irani.
\newblock Separating transparent layers through layer information exchange.
\newblock In {\em European Conference on Computer Vision (ECCV)}, pages
  328--341, 2004.

\bibitem{sarel2005separating}
B.~Sarel and M.~Irani.
\newblock Separating transparent layers of repetitive dynamic behaviors.
\newblock In {\em IEEE International Conference on Computer Vision (ICCV)},
  volume~1, pages 26--32, 2005.

\bibitem{schechner2000separation}
Y.~Y. Schechner, N.~Kiryati, and R.~Basri.
\newblock Separation of transparent layers using focus.
\newblock {\em International Journal of Computer Vision (IJCV)}, 39(1):25--39,
  2000.

\bibitem{schechner2000blind}
Y.~Y. Schechner, N.~Kiryati, and J.~Shamir.
\newblock Blind recovery of transparent and semireflected scenes.
\newblock In {\em IEEE Conference on Computer Vision and Pattern Recognition
  (CVPR)}, volume~1, pages 38--43, 2000.

\bibitem{schechner2000polarization}
Y.~Y. Schechner, J.~Shamir, and N.~Kiryati.
\newblock Polarization and statistical analysis of scenes containing a
  semireflector.
\newblock {\em Journal of the Optical Society of America A (JOSA A)},
  17(2):276--284, 2000.

\bibitem{shih2015reflection}
Y.~Shih, D.~Krishnan, F.~Durand, and W.~T. Freeman.
\newblock Reflection removal using ghosting cues.
\newblock In {\em IEEE Conference on Computer Vision and Pattern Recognition
  (CVPR)}, pages 3193--3201, 2015.

\bibitem{simon2015reflection}
C.~Simon and I.~Kyu~Park.
\newblock Reflection removal for in-vehicle black box videos.
\newblock In {\em IEEE Conference on Computer Vision and Pattern Recognition
  (CVPR)}, pages 4231--4239, 2015.

\bibitem{simonyan2014very}
K.~Simonyan and A.~Zisserman.
\newblock Very deep convolutional networks for large-scale image recognition.
\newblock {\em arXiv preprint arXiv:1409.1556}, 2014.

\bibitem{sinha2012image}
S.~N. Sinha, J.~Kopf, M.~Goesele, D.~Scharstein, and R.~Szeliski.
\newblock Image-based rendering for scenes with reflections.
\newblock {\em ACM Transactions on Graphics (TOG)}, 31(4):100:1--100:10, 2012.

\bibitem{szeliski2000layer}
R.~Szeliski, S.~Avidan, and P.~Anandan.
\newblock Layer extraction from multiple images containing reflections and
  transparency.
\newblock In {\em IEEE Conference on Computer Vision and Pattern Recognition
  (CVPR)}, volume~1, pages 246--253, 2000.

\bibitem{tsin2006stereo}
Y.~Tsin, S.~B. Kang, and R.~Szeliski.
\newblock Stereo matching with linear superposition of layers.
\newblock {\em IEEE Transactions on Pattern Analysis and Machine Intelligence
  (PAMI)}, 28(2):290--301, 2006.

\bibitem{wan2017benchmarking}
R.~Wan, B.~Shi, L.-Y. Duan, A.-H. Tan, and A.~C. Kot.
\newblock Benchmarking single-image reflection removal algorithms.
\newblock In {\em IEEE Conference on Computer Vision and Pattern Recognition
  (CVPR)}, pages 3922--3930, 2017.

\bibitem{wan2017sparsity}
R.~Wan, B.~Shi, A.-H. Tan, and A.~C. Kot.
\newblock Sparsity based reflection removal using external patch search.
\newblock In {\em IEEE International Conference on Multimedia and Expo (ICME)},
  pages 1500--1505, 2017.

\bibitem{xie2012image}
J.~Xie, L.~Xu, and E.~Chen.
\newblock Image denoising and inpainting with deep neural networks.
\newblock In {\em Advances in Neural Information Processing Systems (NIPS)},
  pages 341--349, 2012.

\bibitem{xu2014deep}
L.~Xu, J.~S. Ren, C.~Liu, and J.~Jia.
\newblock Deep convolutional neural network for image deconvolution.
\newblock In {\em Advances in Neural Information Processing Systems (NIPS)},
  pages 1790--1798, 2014.

\bibitem{xue2015computational}
T.~Xue, M.~Rubinstein, C.~Liu, and W.~T. Freeman.
\newblock A computational approach for obstruction-free photography.
\newblock {\em ACM Transactions on Graphics (TOG)}, 34(4):79:1--79:11, 2015.

\bibitem{yang2016robust}
J.~Yang, H.~Li, Y.~Dai, and R.~T. Tan.
\newblock Robust optical flow estimation of double-layer images under
  transparency or reflection.
\newblock In {\em IEEE Conference on Computer Vision and Pattern Recognition
  (CVPR)}, pages 1410--1419, 2016.

\bibitem{yang2016joint}
W.~Yang, R.~T. Tan, J.~Feng, J.~Liu, Z.~Guo, and S.~Yan.
\newblock Joint rain detection and removal from a single image.
\newblock {\em arXiv preprint arXiv:1609.07769}, 2016.

\bibitem{yeung2008extracting}
S.-K. Yeung, T.-P. Wu, and C.-K. Tang.
\newblock Extracting smooth and transparent layers from a single image.
\newblock In {\em IEEE Conference on Computer Vision and Pattern Recognition
  (CVPR)}, pages 1--7, 2008.

\end{thebibliography}
}

\end{document}